# Accuracy Improvement in Differentially Private Logistic Regression: A Pre-training Approach

Mohammad Hoseinpour[1], Milad Hoseinpour[2], Ali Aghagolzadeh[3]

[1]Babol Noshirvani University of Technology, Babol; hpourv@stu.nit.ac.ir
[2]Tarbiat Modares University, Tehran; m.hoseinpour@modares.ac.ir
[3]Babol Noshirvani University of Technology, Babol; aghagol@nit.ac.ir

**Abstract**
Machine learning (ML) models can memorize training datasets. As a result, training ML models on private datasets can lead to the violation of individuals' privacy. Differential privacy (DP) is a rigorous privacy notion to preserve the privacy of the underlying training datasets. However, training ML models in a DP framework usually degrades the accuracy of ML models. This paper aims to increase the accuracy of a DP logistic regression (LR) via a pre-training module. In more detail, we initially pre-train our LR model on a public training dataset without any privacy concern. Then, we fine-tune our DP-LR model with the private dataset. In the numerical results, we show that adding a pre-training module significantly improves the accuracy of the DP-LR model.

## Introduction

ML models are rapidly evolving and expanding their capabilities to unlock a wide range of applications [1]. ML models work by analyzing large amounts of data and adjusting their parameters to accurately capture the complex relationships present in the data. Ideally, we want the parameters of ML models to capture general patterns (e.g., ''patients who smoke are more likely to have heart disease'') instead of specific private details about individual training examples (e.g., "Bob has heart disease") [2]. But a learning algorithm is not oblivious of the private information in a training dataset by default. If we want to use ML for important tasks like creating a heart disease diagnosis model and making it publicly available, we might unintentionally reveal private information about the individuals in the training data. For instance, an adversary with access to this model could attempt a membership inference attack to determine if specific person's data was used in the training dataset of the model [3]. Also, the adversary could build on reconstruction attacks to extract private training data [4]. DP is a framework that uses strong mathematical guarantees to effectively handle the risk of privacy breaches [5]. Furthermore, DP enables us to quantitatively measure the degree of privacy guarantee in a computation. The underlying idea of this approach is to embed carefully calibrated random noise in a computation [6]. DP guarantees that the noisy outputs of the computation do not disclose the private information of individuals in the input data. However, acheiving DP in computations is associated with a reduction in accuracy. Specifically, DP-ML models usually involve a trade-off between the degree of privacy afforded to individuals in the training dataset and the accuracy of the model. In other words, to maintain a meaningful level of privacy protection, the model's accuracy tends to decrease significantly.

There is a rich and growing literature on DP-ML models. In this context, the inherent trade-off between accuracy and privacy in DP-ML models is a challenge that necessitates further consideration. This paper aims to propose an approach for boosting the accuracy of DP-LR models based on the publicly available datasets. The main contribution of this paper is to develop a new architecture for DP-LR models by adding a pre-training module. This module derives a general pattern from the publicly available dataset in the specific context of an LR task. This general pattern makes the LR model more robust against the required randomization of DP. Thus, this pre-training step would help us to amplify the accuracy of our model for a given privacy level.

The rest of this paper is as follows. In the background section, the required preliminaries are presented. We propose our pre-training approach for boosting the accuracy of a DP-LR model in the third section. In the fourth section, we provide numerical results for reflecting the effectiveness of pre-training for accuracy improvement. Following that, the conclusion is presented.

## Background

In this section, we briefly recall the definition of DP and overview the basic concepts of LR.

### A. Differential Privacy

DP is the de-facto standard for privacy protection, which gives us a framework to quantitatively reason about privacy [7], [8]. DP ensures that the output of a computation will be roughly unchanged whether or not an individual's data is used, so restricting an adversary's ability to deduce information about specific data points related to individuals [9]. The main idea for satisfying this condition is to perturb the computation by injecting a calibrated amount of noise to mask the contribution of each individual in the dataset. In the following, we present the formal definition of DP and a couple of remarks about this notion.

**Definition 1** (Differential Privacy). For $\epsilon > 0$ and $\delta \in [0,1]$, a randomized algorithm $M : X^n \to R$ is $(\epsilon, \delta)$-differentially private if for every pair of neighboring datasets $x \sim x' \in X^n$ (i.e., $x$ and $x'$ differ in one element) and for any subset of the output space $S \subseteq R$, the following holds [6]:

$$\Pr[M(x) \in S] \leq e^\epsilon \Pr[M(x') \in S] + \delta \quad (1)$$

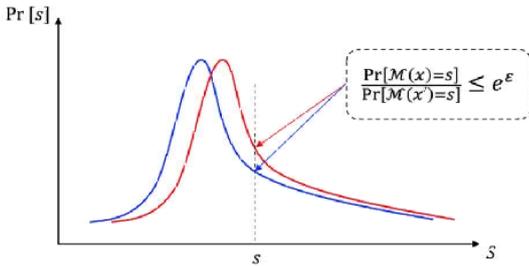

Figure 1. Closeness of the output distributions of a DP mechanism over neighboring datasets.

Figure 1 shows a high-level view of the performance of a DP mechanism. It shows that the chance of every event $s$ is almost the same in any two neighboring datasets $x \sim x' \in X^n$. The aforementioned definition guarantees that no individual's data has a large impact on the output of M. More formally, when an $(\epsilon, \delta)$-differentially private algorithm runs on two neighboring datasets, the resulting distributions over the output space will be very similar, and this similarity is captured by a multiplicative factor $e^\epsilon$. The original definition of DP does not include the additive term $\delta$. We use the variant introduced by Dwork [6], which allows for the possibility that plain $\epsilon$-DP is broken with probability $\delta$ (which is preferably smaller than $1/|x|$). The required noise for satisfying DP is calibrated based on the global sensitivity of the computation. We formalize the mathematical definition of the global sensitivity in the following [8].

**Definition 2** (Global Sensitivity). For a function $f : X^n \to R^k$, the global sensitivity over all pairs of neighboring datasets $x \sim x' \in X^n$ is

$$GS(f) = \max_{x \sim x' \in X^n} \|f(x) - f(x')\|_2 \quad (2)$$

where $\|\cdot\|_2$ is the $l_2$- norm [5].

To attain DP in a computation, it is necessary to limit the impact of each individual data point on that computation. Hence, we should implement an appropriate privacy mechanism for perturbing the computation, e.g., Laplace mechanism and Gaussian mechanism, known as the additive noise approaches. Moreover, perturbation techniques for satisfying DP in a computation are mainly classified into two basic categories: (1) adding calibrated random noise to the input data, and (2) adding calibrated random noise to the outputs [8], [10].

**Definition 3** (Gaussian Mechanism). If $f : \mathcal{X}^n \to \mathbb{R}^k$, then Gaussian mechanism directly add Gaussian noise to the output of the computation in the following sense:

$$\mathcal{M}(x) = f(x) + (Y_1, \cdots, Y_k). \quad (3)$$

Where $(Y_i)_{i=1}^k$ are independent random variables drawn from [11]:

$$\mathcal{N}(0, 2\ln(1.25/\delta)\Delta(f)^2/\epsilon^2) \quad (4)$$

*B. Logistic Regression*

Logistic regression is a supervised learning algorithm that efficiently could categorizes a new data point by estimating the probability of belonging to a particular class. While logistic regression can be expanded to handle more than two categories, it is frequently utilized for binary classification tasks. In this paper, our focus will be specifically on binary logistic regression.

The typical setup for logistic regression is as follows. Given the provided input feature vector $X = (x_1, x_2, \ldots, x_k)$ for a data point, there is an outcome $y$ that falls into one of two categories (0 or 1). Also, the probability of label y for a specific category can be calculated by the following equation:

$$P(y = 1|X) = sigmoid(z) = \frac{1}{1 + e^{-z}} \quad (5)$$

where
$$z = wx + \beta, \quad (6)$$

where $w$ is a $d$-dimensional vector of parameters, $x \in \mathbb{R}^d$ is the feature vector, and $\beta$ is a real number.

The equation (6) looks familiar from linear regression. This equation is transformed by the Sigmoid function. In Sigmoid function, the values fall between 0 and 1, and can be interpreted as probabilities. This resulting probability is then compared to a threshold (usually 0.5) to predict a class for y based on X.

When fitting our model, the goal is to find the parameters $w$ that minimizes the loss function that defines how well the model is performing. Simply, the goal is to make predictions as close to 1 when the outcome is 1 and as close to 0 when the outcome is 0. Indeed, the loss function reflects how well our model fits the data. A suitable loss function in logistic regression is called the Log-Loss, or binary cross-entropy:

$$J = \sum_{i=1}^m -(y_i\log(p_i) - (1 - y_i)\log(1 - p_i)), \quad (7)$$

where $m$ is the number of samples, indexed by $i$, $y_i$ is the true class for the index $i$, and $p_i$ is the model prediction for the index $i$ [12].

*C. Gradient Descent*

Gradient descent (GD) is a straightforward iterative optimization algorithm used to estimate a set of coefficients that leads to the minimum value of a convex function. Simply, it will find suitable coefficients for our logistic regression model that minimizes the loss function. We assume that we have some convex function representing the error of our ML algorithm. During the iterative process, GD continually updates the coefficients of our model by moving towards the minimum of our loss function. In our case, the model takes the form (5) and the error function takes the form (7). Our objective is to discover the optimal values for the coefficients $w$ and $\beta$ that minimize the loss function. To achieve this, we will utilize the gradient, which indicates both the direction and the rate at which the function is increasing. As we aim to locate the minimum of this function, we can progress in the opposite direction of its increasing trend. To achieve this, we first compute the derivatives of our loss function $J$ from (7) with respect to each coefficient that needs updating. We then perform iterative updates until convergence is reached:

$$w = w - \alpha \frac{\partial}{\partial w} J \tag{8}$$

$$\beta = \beta - \alpha \frac{\partial}{\partial \beta} J \tag{9}$$

In the following you can see the algorithm of the gradient descent [13]:

---
**Algorithm 1**: Gradient Descent

Inputs: *noise parameter* ($\sigma>0$), Learning Rate $\alpha$,
1: $w_0$ = initial value for w
2: for t=1,2,…,T :
3:   $g_t = \nabla J(w_{t-1})$;
4:   $w_t = w_{t-1} - \eta g_t$;
5: Return $w_T$

---

**The Model**

*A. Noisy Gradient Descent*

In our model, we embed the required randomization in the training process of the logistic regression model, which is based on the GD algorithm. That is, we add a calibrated amount of Gaussian noise to each iteration of the GD updating rule. For calibrating the required amount of Gaussian noise, we need to compute the global sensitivity of step 5 in Algorithm 1. To do so, we only focus on $g_t = \nabla J$ which touches the input datasets:

$$GS(\nabla J(x)) = \max_{x \sim x'} \|\nabla J(x) - \nabla J(x')\|_2. \tag{10}$$

In general, there is no upper-bound for $GS(\nabla J(x))$. In this regard, we implement the gradient clipping approach to define an upper-bound for the $GS(\nabla J(x))$ [14]. However, due to the robustness of the GD algorithm, we can clip the gradient $g_t = \nabla J$ in each iteration of the algorithm with an arbitrary bound $C$. Thus, the gradient vector should be replaced with $g_t/max(1, \|g_t\|_2/C)$, where $C$ is the clipping threshold. In this regard, $GS(\nabla J(x))$ is equal to

$$GS(\nabla J(x)) = \max_{x \sim x'} \|\nabla J(x) - \nabla J(x')\|_2 \leq \max_{x \sim x'} (\|\nabla J(x)\|_2 - \|\nabla J(x')\|_2) = 2C. \tag{11}$$

Thus, for achieving $(\epsilon', \delta') - DP$ in each iteration of GD, we should add Gaussian noise with parameter $\sigma \geq \frac{2C}{n\epsilon'}\sqrt{2\ln\left(\frac{1.25}{\delta'}\right)}$ to $g_t = \nabla J$ in each iteration. In the following you can see the algorithm of the noisy gradient descent [13]. You can see the gradient clipping and perturbation with zero mean Gaussian noise $Z_t$ in Figure 2.

---
**Algorithm 2**: Noisy Gradient Descent

Inputs: *noise parameter* ($\sigma>0$), Learning Rate $\alpha$,
1: $w_0$ = initial value for w
2: for t=1,2,…,T :
3:   $g_t = \nabla J(w_{t-1})$;
4:   clip the gradient:
$$g_t^{clip} = \frac{g_t}{max(1, \|g_t\|_2/C)}$$

5:   $g'_t = g_t^{clip} + N(0, \sigma^2 I_d)$;
6:   $w_t = w_{t-1} - \alpha g'_t$;
7: Return $w_T$

---

*B. Pre-training Module*

One of the main challenges in training a DP-LR model is the utility and privacy trade-off. That is, for having a high level of privacy protection the accuracy of the LR model is degrade, and vice-versa. To address this challenge, due to Figure 3, we add a pre-training module to the logistic regression model. Specifically, we pre-trained our model on a public training dataset that there is no privacy concern about it. Subsequently, we fine-tune our model via the DP-LR with the private dataset.

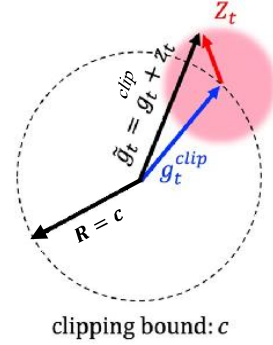

Figure 2. A high-level view of gradient clipping and perturbation procedure

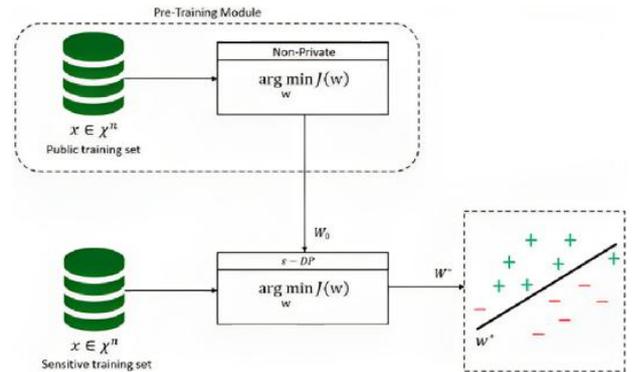

Figure 3. A high-level view of the proposed approach for accuracy improvement via the pre-training module

**Numerical Results**

In this section, we first reflect the inherent trade-off between the accuracy and privacy guarantee of the DP-LR model. Then, we demonstrate the boosting effect of a pre-training module for the accuracy of the DP logistic regression. For the purpose of this paper, we deploy a privacy-sensitive dataset, where the individuals participating in the training dataset are concerned with the leakage of their sensitive information. Specifically, we have a healthcare training dataset with a size of 400. this training dataset has two medical features which are used to classify patients diagnosed with heart disease as high risk and low risk. For boosting the accuracy of our DP-LR model, we implement a publicly available dataset with the same features and labels with a size of 400 to use in our pre-training module. We should mention that the proposed model could be used in settings with rich feature space with multi classes for categorization.

### A. Differentially Private Logistic Regression

We calculate the accuracy of DP-LR model in different privacy regimes. Due to Table 1, there is a trade-off between $\varepsilon$ and accuracy of the model with and without pre-training module. In fact, you can see the accuracy of the DP-LR model decreases as the privacy level increases.

Figure 4 shows the loss function of the non-private LR model. As You can see the loss function is strictly descending and smooth. However, as Figure 5 shows, the loss function in the DP-LR is rippled due to the noise addition in the updating rule of the gradient descent.

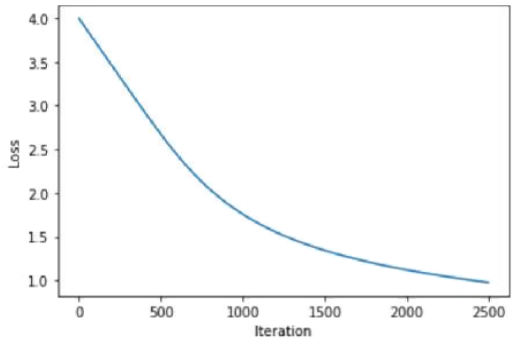

Figure 4. Loss versus iteration of the LR model with almost no privacy constraint (ε=100) and training accuracy 67.50%

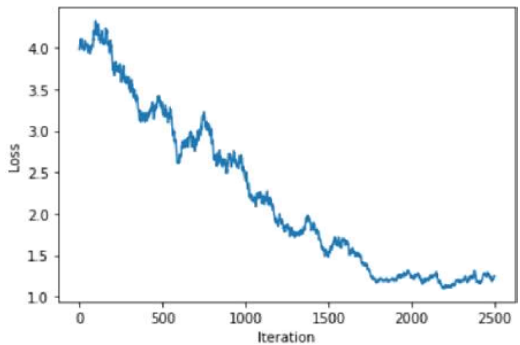

Figure 5. Loss versus iteration graph of the DP-LR model with ε=1 and 60.25% training accuracy

### B. Pre-Trained Differentially Private Logistic Regression

In this part, we calculate the accuracy of the pre-trained DP-LR model in different privacy regimes. As you can see in Table 1, the pre-trained model has a better performance for a given privacy level. You can see the comparison of the model accuracy with and without the pre-training module in Table 1.

Figure 6 shows the loss function of the DP-LR model with the pre-training module. You can see that the level of loss reduces significantly via pre-training in comparison to Figure 5.

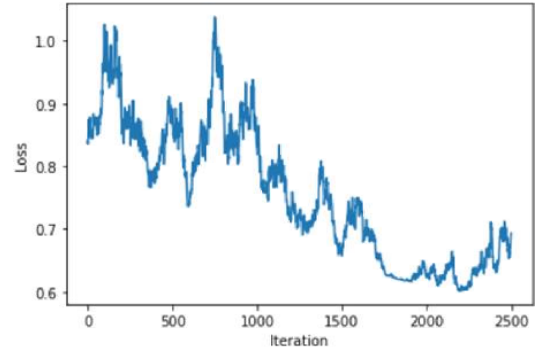

Figure 6. Loss versus iteration graph of a pre-trained DP-LR model with ε=1 and training accuracy 70.50%

Due to Table 1 the boosting effect of the pre-training module is negligible in high privacy regimes. For example, when $\varepsilon = 0.05$ the accuracy improvement is %0.5, which is much lower than the accuracy improvement for $\varepsilon = 0.5$, %7.5. Also, Figure 7 shows the accuracy of the DP-LR model with pre-training module versus the standard deviation.

Table 1. The accuracy of DP-LR model with and without pre-training module

| ε | Accuracy With No Pre-training Module | Accuracy With Pre-training Module | Enhancement |
|---|---|---|---|
| 0.01 | %29.75 | %29.75 | ≈0 |
| 0.05 | %33.00 | %33.50 | %0.5 |
| 0.1 | %40.25 | %41.25 | %1.25 |
| 0.5 | %53.25 | %60.00 | %7.25 |
| 1 | %60.25 | %70.50 | %10.25 |
| 5 | %66.25 | %77.25 | %11 |
| 10 | %66.50 | %77.50 | %11 |
| 15 | %67.00 | %77.50 | %10.50 |
| 150 | %67.50 | %78.00 | %11.50 |

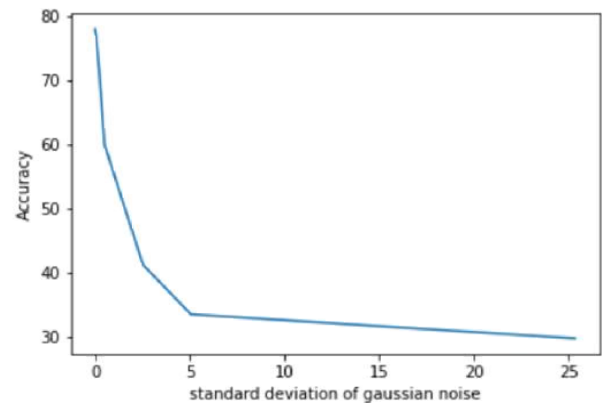

Figure 7. Accuracy of DP-LR model with pre-training module versus the scale of the noise ($\sigma$)

### C. Impact of Noisy Gradient Descent on Decision Boundary

In Figure 8 – Figure 11, you can see how adding noise in the GD algorithm impacts on the decision boundary of pre-trained DP-LR.

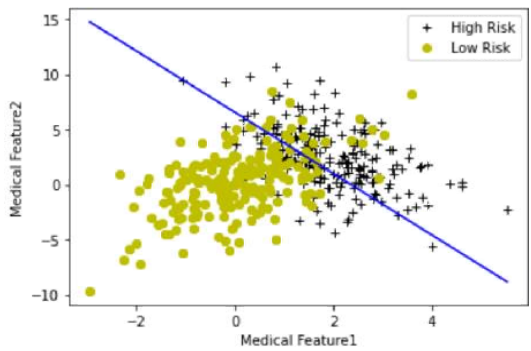

Figure 8. Decision boundary of the pre-trained DP-LR model with no privacy constraint and 78.00% training accuracy

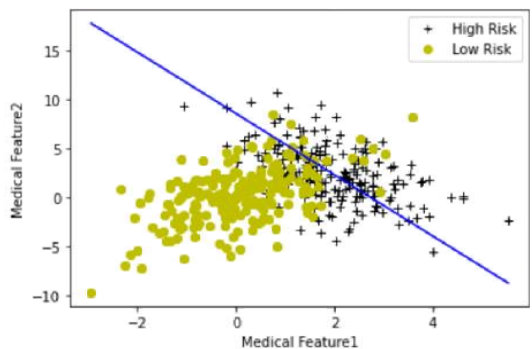

Figure 9. Decision boundary of the pre-trained DP-LR model with ε=1 and 70.50% training accuracy

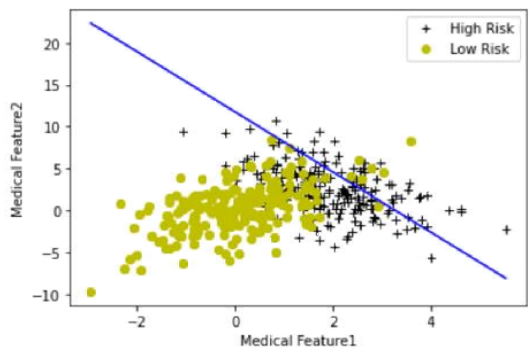

Figure 10. Decision boundary of the pre-trained DP-LR model with ε=0.5 and 60.00% training accuracy

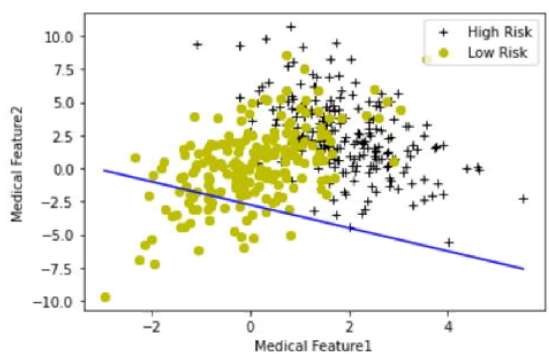

Figure 11. Decision boundary of the pre-trained DP-LR model with ε=0.1 and 41.25% training accuracy

**Conclusion**

In this paper, we proposed an approach to improve accuracy in a DP-LR model. To do this, we first pre-trained our logistic regression model on a public dataset, which has no privacy concern. Then, we fine-tuned our model by a sensitive dataset under the DP constraint. Numerical results showed adding a pre-training module can significantly increase the accuracy of our model. For instance, when $\varepsilon = 1$, the accuracy of the DP-LR model increases by about 10 percent.

**Acknowledgement**

We would like to thank Matin Nassabian for his very helpful suggestions and comments on this paper.